\documentclass{article}
\usepackage[numbers]{natbib}
\usepackage[final]{neurips_2023_arxiv} %

\usepackage[utf8]{inputenc} %
\usepackage[T1]{fontenc}    %
\usepackage{hyperref}       %
\usepackage{url}            %
\usepackage{booktabs}       %
\usepackage{amsfonts}       %
\usepackage{nicefrac}       %
\usepackage{microtype}      %
\usepackage[dvipsnames]{xcolor}
\usepackage{bm}
\usepackage{courier}
\usepackage{soul}
\usepackage{graphicx}
\usepackage{xspace}
\usepackage{subcaption}
\usepackage{colortbl}
\usepackage{flafter}
\usepackage[utf8]{inputenc}
\usepackage[most]{tcolorbox}

\providetoggle{showcomments}
\settoggle{showcomments}{true}

\iftoggle{showcomments}{
    \newcommand{\resolved}[3][]{\ifstrequal{#1}{resolved}{\textcolor{blue}{RESOLVED:}~\textbf{{\MakeUppercase #2:}}~{#3}}{\textbf{\MakeUppercase #2:}~#3}}
    \newcommand{\lisa}[2][]{\textcolor{ForestGreen}{\resolved[#1]{lisa}{#2}}}
    \newcommand{\fantine}[2][]{\textcolor{red}{\resolved[#1]{fantine}{#2}}}
    \newcommand{\jasper}[2][]{\textcolor{violet}{\resolved[#1]{jasper}{#2}}}
    \newcommand{\jrru}[2][]{\jasper[#1]{#2}}
    \newcommand{\tm}[2][]{\textcolor{magenta}{\resolved[#1]{thomas}{#2}}}
    \newcommand{\lluis}[2][]{\textcolor{RedOrange}{\resolved[#1]{lluis}{#2}}}
    \newcommand{\mostafa}[2][]{\textcolor{MidnightBlue}{\resolved[#1]{mostafa}{#2}}}
    
    \newcommand{\todo}[1]{\textcolor{blue}{\textbf{TODO:} #1}}
}{
    
    \newcommand{\todo}[1]{}
    \newcommand{\lisa}[2][]{}
    \newcommand{\fantine}[2][]{}
    \newcommand{\jasper}[2][]{}
    \newcommand{\jrru}[2][]{}
    \newcommand{\tm}[2][]{}
    \newcommand{\lluis}[2][]{}
    \newcommand{\mostafa}[2][]{}
    
}

\newcommand*{\eg}{\textit{e.g.}\@\xspace}
\newcommand*{\ie}{\textit{i.e.}\@\xspace}

\newcommand{\para}[1]{\par\noindent\textbf{#1}\quad}

\newtcolorbox[auto counter]{promptfloat}[2][]{%
    float=b!,%
    blend before title=dash hang,%
    title={\textbf{Prompt~\thetcbcounter:} #2},%
    colback=yellow!10!white,%
    colframe=blue!35!black,%
    #1}

\newcommand{\prompt}[3]{%
    \begin{promptfloat}[label={#1}]{#2}
    {\footnotesize \texttt{#3}}
\end{promptfloat}
}

\newtcolorbox{promptcontinuesfloat}[1][]{%
    float=t!,%
    blend before title=dash hang,%
    colback=yellow!10!white,%
    colframe=blue!35!black,%
    #1}

\newcommand{\promptcontinues}[1]{%
    \begin{promptcontinuesfloat}
    {\footnotesize \texttt{#1}}
\end{promptcontinuesfloat}
}

\title{How (not) to ensemble LVLMs for VQA}
\author{%
  Lisa Alazraki\thanks{Work done during an internship at Google.\newline\indent{}\quad Contact:
  \texttt{lisa.alazraki20@imperial.ac.uk}, or
  \texttt{\{jrru,mensink\}@google.com},
  }\\
  Imperial College London
  \And
  Lluis Castrejon\\
  Google Research
  \And
  Mostafa Dehghani\\
  Google DeepMind
  \And
  Fantine Huot\\
  Google DeepMind
  \AND
  Jasper Uijlings\\
  Google Research
  \And
  Thomas Mensink\\
  Google Research
}

\begin{document}

\maketitle

\begin{abstract}
This paper studies ensembling in the era of Large Vision-Language Models (LVLMs). Ensembling is a classical method to combine different models to get increased performance. In the recent work on Encyclopedic-VQA~\cite{mensink2023encyclopedic} the authors examine a wide variety of models to solve their task: from vanilla LVLMs, to models including the caption as extra context, to models augmented with Lens-based retrieval of Wikipedia pages. Intuitively these models are highly complementary, which should make them ideal for ensembling. Indeed, an oracle experiment (Fig.~\ref{oracleFigure}) shows potential gains from 48.8\% accuracy (the best single model) all the way up to 67\% (best possible ensemble). So it is a trivial exercise to create an ensemble with substantial real gains. Or is it?

\end{abstract}

\section{Introduction}
\begin{figure}[t]
    \centering
    \vspace{-5mm}
    \includegraphics[width=\linewidth]{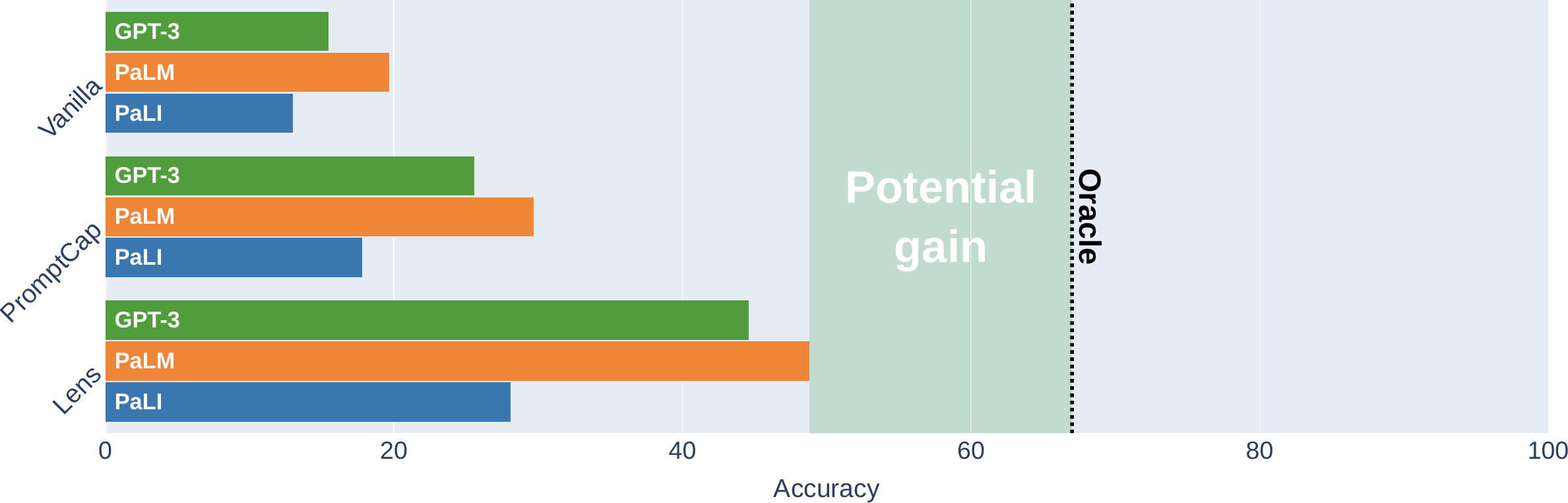}
    \vspace{-5mm}
    \caption{\small{Encyclopedic-VQA baselines and oracle ensemble of nine LVLMs. All results are on the single-hop single-answer questions following the main experiments in~\cite{mensink2023encyclopedic}.}}
    \vspace{-2mm}
    \label{oracleFigure}
\end{figure}

\begin{figure}[t]
    \centering \includegraphics[width=\linewidth]{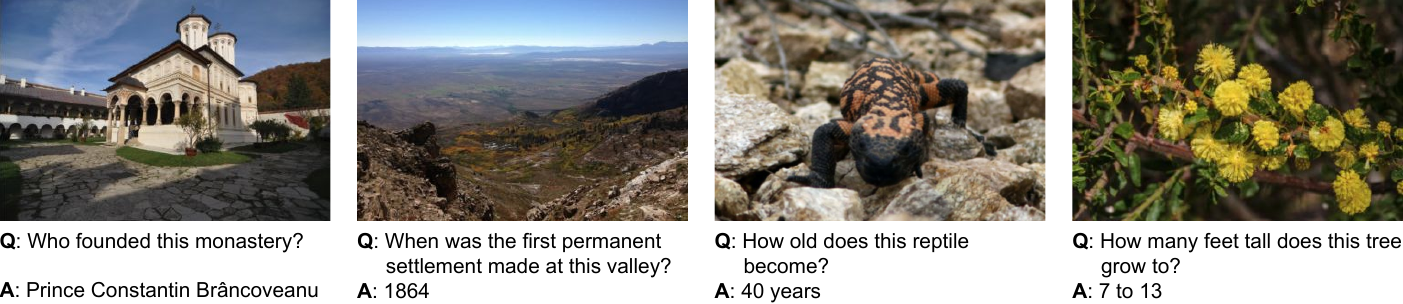}
    \caption{\small{Examples from the Encyclopedic-VQA task.}}
    \vspace{-5mm}
    \label{encvqaFigure}
\end{figure}

Large Vision-Language models (LVLMs) have achieved impressive results on Visual Question Answering (VQA)~\cite{alayrac22neurips,chen22arxiv,hao22arxiv,yang22aaai}. Ensembling multiple LVLMs has the potential to increase performance even further. In this work, we focus on Encyclopedic-VQA -- a recently-introduced VQA task \cite{mensink2023encyclopedic} asking questions about detailed properties of finegrained categories (Fig.~\ref{encvqaFigure}). In \cite{mensink2023encyclopedic} they use nine LVLMs: PaLI \cite{chen22arxiv}, PaLM \cite{chowdhery2022} and GPT-3 \cite{brown2020language}, each deployed with three different augmentation strategies: (1) PromptCap \cite{hu2022promptcap}, (2) Wikipedia sections retrieved via Google Lens \cite{googlelens} and (3) without any augmentation (‘vanilla'). Intuitively, these approaches are quite different and complementary, which makes them ideal for ensembling from a theoretical perspective~\cite{dietterich02ensemble,yang21ensemble}. Therefore in this paper we set out to create a strong ensemble out of these models.

To show the potential of ensembling, we carry out an oracle experiment on Encyclopedic-VQA (Fig.~\ref{oracleFigure}), on the single-hop, single-answer questions of the dataset~\cite{mensink2023encyclopedic}. In particular, we select for each VQA example the best answer out of those given by the nine LVLMs. Whereas the best single model achieves 48.8\% accuracy, the best possible ensemble achieves 67.0\% accuracy. This demonstrates we have a large potential gain of 18.2\%!

Hence we explore in this paper a variety of ensembling methods: classical ensembling techniques such as majority voting and using model confidence (Sec.~\ref{sec:classical}), prompting the LVLMs to do self-reflection (Sec.~\ref{sec:self-reflection}), and finally using an external evaluator model to judge which answer is correct (Sec.~\ref{sec:external_eval}).
In our exploration we aim to find which ensembling techniques work, which do not, and why.

Our contributions are the following:
(1) we identify the large potential gain of ensembling multiple LVLMs on Encyclopedic-VQA;
(2) we explore a variety of ensembling methods for LVLMs: classical methods, self-reflection, and external evaluation by another LVLM;
(3) we increase performance on Encyclopedic-VQA by 4.6\% through a model cascade using external evaluation. Our analysis shows that the majority of this boost is attributed to the utilization of a larger model for evaluation and a singular critical observation;
(4) most of the potential gain remains untapped and most of our ensemble strategies were not as successful as envisioned. This leads us to conclude that effectively ensembling LVLMs is challenging.
\section{Classical Ensembling Methods}\label{sec:classical}

We begin our investigation into ensembling from established methods that have been widely used before: majority voting \cite{58871, oniani2023large, imani2023diversigate} and using the model's own confidence \cite{LI20143120, Gitman_2023, rosales2023evaluation}. Before doing so, we detail our oracle experiment.

\paragraph{Oracle Ensembling}
We run inference on Encyclopedic-VQA using the nine LVLMs in \cite{mensink2023encyclopedic}. 
The oracle selector picks one of the LVLMs yielding the correct answer, or a random one if none of the LVLMs has produced the correct answer.
We evaluate the oracle selector on the test set and obtain 67.0\% BEM accuracy \cite{bulian2022tomayto}. The oracle thus shows a large potential improvement of 18.2\% over the accuracy achieved by PaLM with Lens (48.8\%), the best-performing single LVLM in~\cite{mensink2023encyclopedic}.

\subsection{Majority Voting}
In majority voting, we use the most voted answer among the outputs of the nine LVLMs. 
We adopt BEM-style soft-matching (\ie using BERT matching \cite{bulian2022tomayto} to determine whether two answers are the same). We evaluate majority voting on the Encyclopedic-VQA test set and obtain 45.3\%, much lower than the best single model (48.8\%). 
Hence we conclude that majority voting does not improve the VQA performance for this task.

\subsection{Model Confidence}

\begin{table}[!t]
    \vspace{-.9cm}
    \caption{\small{Results on the Lens retrieval setup.}}
    \centering
    \begin{subtable}{0.5\linewidth}
        \centering
        \resizebox{!}{15mm}{
        \begin{tabular}{lc}
            \toprule
            \textbf{} & \textbf{{Acc. (BEM)}}              \\
            \cmidrule{2-2}
            {PaLM w/ Lens (baseline)} & {49.4\%} \\
            \midrule
            {Choosing max prob. answer} & {49.2\%} \\
            {Weighted voting} & {49.2\%} \\
            {Logistic regression} & {50.6\%} \\
            \midrule
            {Oracle ensemble} & {57.5\%} \\
            \bottomrule
        \end{tabular}        
        }\vspace{-1mm}
        \caption{Classical ensembling results.}
        \label{tab:sequenceProbs}        
    \end{subtable}%
    ~
    \begin{subtable}{0.5\linewidth}
        \centering      
        \resizebox{!}{15mm}{
        \begin{tabular}{lccc}
        \toprule
        & PaLI & PaLM & GPT-3\\
        \cmidrule{2-4}
        \textbf{BEM} ($\uparrow$) & 34.6\% & 49.4\% & 47.1\%\\
        \midrule
        \textbf{ECE} ($\downarrow$) & 0.121 & 0.175 & 0.272 \\
        \textbf{Brier} ($\downarrow$) & 0.182 & 0.272 & 0.278 \\
        \midrule
        \textbf{ECE}-re ($\downarrow$) & 0.039 & 0.131 & 0.127 \\
        \textbf{Brier}-re ($\downarrow$) & 0.164 & 0.256 & 0.221 \\
        \bottomrule
        \end{tabular}
        }\vspace{-1mm}
        \caption{Calibration metrics.}
        \label{tab:calibration_metrics}
    \end{subtable}   
    \vspace{-5mm}
\end{table}

In this section we focus on using the sequence probabilities of the LVLMs as a signal for ensembling. Sequence probabilities have been used to estimate confidence in a QA answer in \cite{si2022prompting}. Similarly to \cite{si2022prompting}, we normalize the probabilities for sequence length, \ie 
$\bar{p} = p^\frac{1}{N}$, with $p=\prod_{i=0}^{N} \ p(t_i|t_{<i})$ for a sequence of $N$ tokens.

\para{Ensemble methods.}
For the following experiments we use three LVLMs: PaLI, PaLM, and GPT-3 all with the Lens retrieval setup, for a subset of 1,000 Encyclopedic-VQA examples. 
We attempt three different strategies to ensemble these models according to their (normalized) probabilities:
\begin{enumerate}
    \item \textbf{Max probability.} Choose per example the LVLM with the highest probability. 
    \item \textbf{Weighted voting.}
    If the same answer is given by multiple LVLMs use the average probability across the LVLMs, otherwise the probability of the LVLM producing the answer.
    \item \textbf{Classification.} Train a classifier to learn the weights for weighted voting, using the probabilities of all three models as inputs.
\end{enumerate}

The resulting BEM scores (Tab.~\ref{tab:sequenceProbs}), show that all strategies
perform similarly to the baseline. The logistic regression classifier slightly outperforms it (by 1.2\%), yet its BEM score is far below that achieved by the oracle ensemble of those three models.
We also try adding to the classifier feature transformations (such as Z-score normalisation, sqrt, or power), feature combinations (\eg $p1 \times p2$), and hidden layers in an MLP to add more expressivity, all without much results.
We hence conclude that LVLM sequence probabilities are not reliable enough to be used for ensembling.
In the next paragraph we discuss if this could be due to miscalibration.

\para{Calibration.}
In order to understand if the sequence probabilities of the different LVLMs are comparable, we investigate here whether these are calibrated w.r.t. VQA accuracy. In other words, we examine whether these generative probabilities are also representative of the likelihood of an answer being correct. Perfect calibration would mean that if the 
probability is 80\%, we find (empirically) that the answer is also correct 80\% of the times.

To measure calibration we use the ECE \cite{Naeini_Cooper_Hauskrecht_2015} and the Brier score~\cite{Brier1950VERIFICATIONOF} as metrics, which we evaluate over the same subset of examples and LVLMs as above.
The results are in Tab.~\ref{tab:calibration_metrics}.
We observe (1) that the accuracy of the chosen examples is slightly above the average; (2) that the Brier score and the ECE metric are correlated, \ie that the method with the lowest Brier score also has the lowest ECE, (3) that the PaLI LVLM is best calibrated, but does not yield the highest BEM score.

We also optimize the ECE/Brier scores by temperature scaling, \ie we derive $p$ from the sequence log-likelihood $l$ as $p = \exp{l/t}$, where the temperature $t$ is individually tuned for each LVLM (see Appendix \ref{sec:calibration}). We find that this method of calibration significantly improves the ECE and Brier scores as evidenced in Tab.~\ref{tab:calibration_metrics}. 
However, ensembling via model selection using the maximum of the calibrated probabilities performs 
identically to the non-calibrated probabilities.
We hence conclude that while temperature scaling leads to better calibrated LVLMs, their sequence probabilities remain a weak signal for ensembling.

\para{Limitations.}
In this section we use sequence probabilities as a measure of the model's confidence in VQA answers, following \cite{si2022prompting}. The relatively low ECE and Brier scores justify this approach. However, recent work \cite{kuhn2023} criticizes this method, as different yet semantically equivalent answers would yield different probabilities despite being equally correct. Future work could investigate the use of semantic likelihood \cite{kuhn2023} in place of sequence probability.

Another limitation is that for typical weighted voting methods one averages the models over each class. For LVLMs something similar could be achieved by first collecting all answers, and then have each LVLM output the probability of each of these answers. Unfortunately, due to the limitations of the APIs this was not possible to do for all models and we could not explore this avenue.
\section{Self-Reflection}\label{sec:self-reflection}

\begin{figure}[!t]
    \centering
    \includegraphics[width=\textwidth]{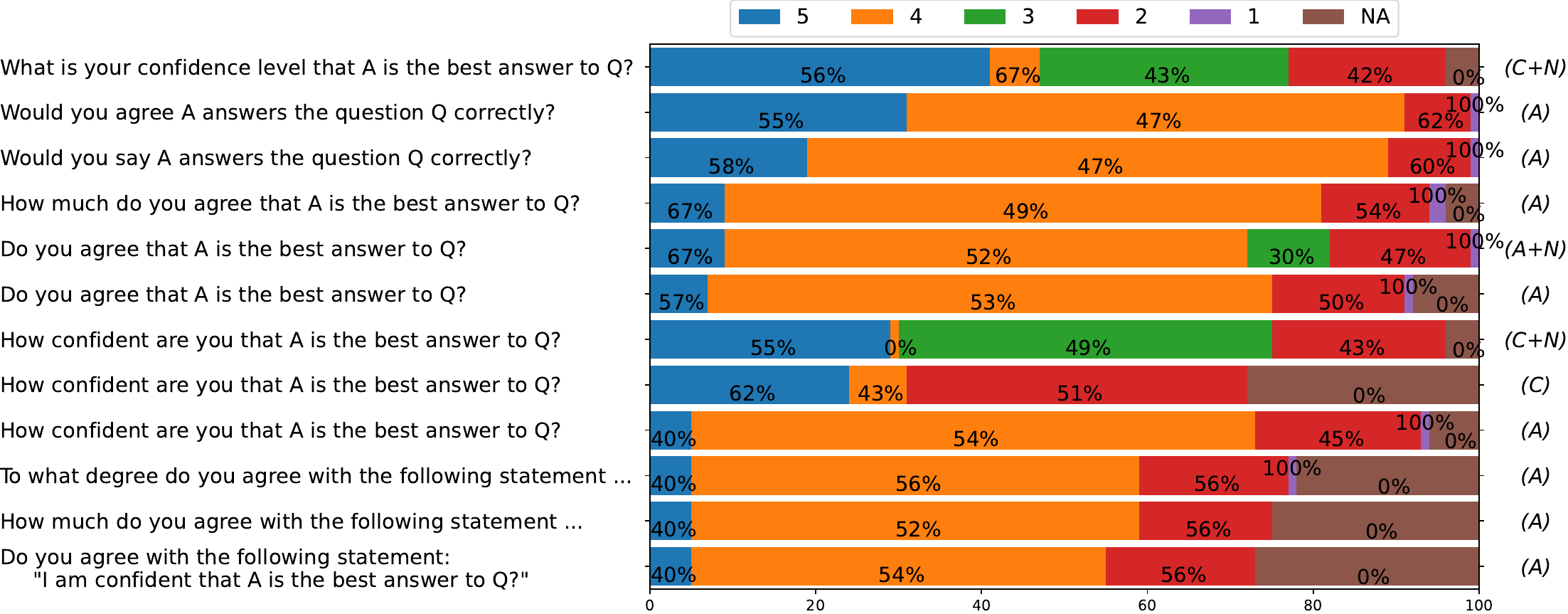}
    \caption{\small{Variations of the same prompt to elicit a Likert-Scale confidence prediction, evaluated using PaLM on 100 examples. The width of the bars shows the proportion of examples falling in each scale. The percentage represents the BEM accuracy of the questions in each Likert-bucket, which ideally should be high when the model is very confident abouts its answer (scale 5 in blue) and low when the model is very unconfident about its answer (scale 1 in purple). `C' means our Likert scale is phrased in confidence: `very confident', `confident', `neither confident nor unconfident', `unconfident', `very unconfident'. `A' indicates the same scale but in terms of `agree'. For `N' we rephrased the middle option (in green) as `neutral'. }}

    \label{fig:likert_scale_palm}
    \vspace{-4mm}
\end{figure}
\begin{figure}[t]
    \centering
    \includegraphics[width=.99\textwidth]{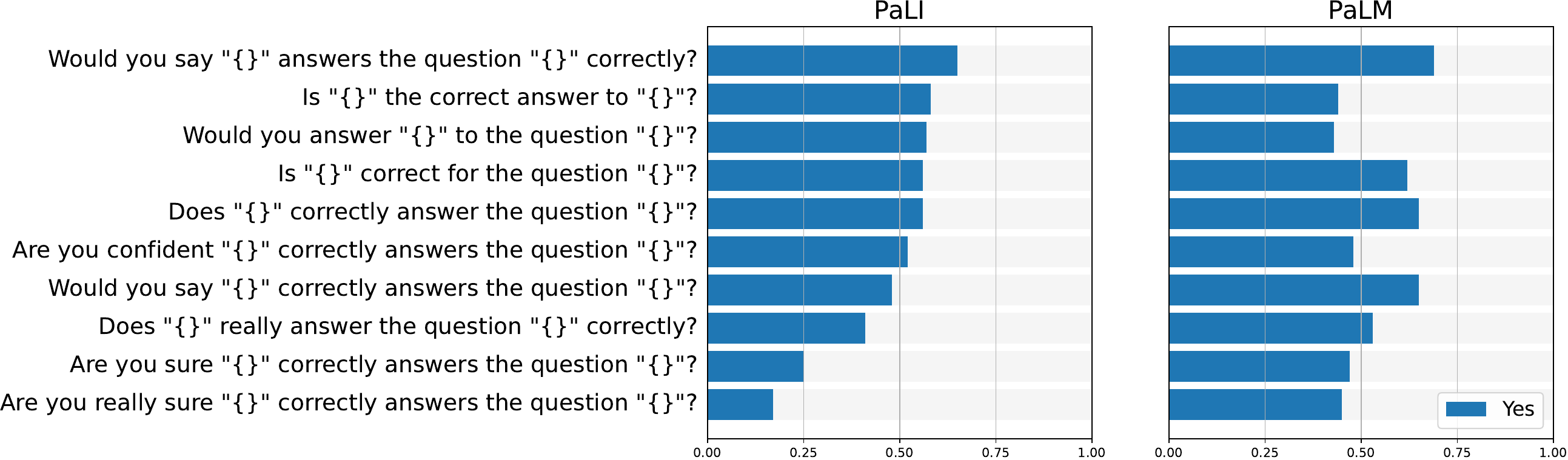}
    \caption{\small{Evaluation of different prompts to get a yes/no confidence reflection. Ten different prompts are evaluated for PaLI and PaLM on the same subset of 100 examples. We observe a large variation in the distribution of the yes/no responses between the different prompts, and across the two models.
    }}
    \label{fig:prompt_variations}
    \vspace{-4mm}
\end{figure}

LVLMs open up a new and interesting way of obtaining a confidence measure by having them self-reflect on their own prediction through prompting~\cite{tian2023just, xiong2023llms}. We explore this here.

In preliminary experiments we tried prompting the LVLMs to output a numerical value for their confidence, as previously explored in \cite{xiong2023llms}. These experiments showed that these models are not good at providing a numerical value for complex concepts such as confidence or correctness. Hence we focus on language variations of confidence estimations, similarly to \cite{tian2023just}, using yes/no questions and the 5-point Likert scale.

\para{Likert scale.}
Prompts including a 5-point Likert scale are too long for PaLI, so we perform these experiments using PaLM only. Fig.~\ref{fig:likert_scale_palm} shows several variations of the same prompt to elicit a confidence score. In this figure, the size of the bars corresponds to the percentage of answers which fall in a confidence bucket: 5 (blue) for very confident, and 1 (purple) for very unconfident. We also plot the corresponding BEM accuracy scores averaged over the answers in each bucket as a percentage on top of it. Now examining the first prompt (top), we see that the BEM accuracy is 56\% for answers where the LVLM is `very confident'. For answers where it is just `confident', the BEM accuracy increases to 67\% whereas we would expect a decrease if the self-reflection was good. Results are even more dramatic for the second row: BEM accuracy is highest (62\%) for answers which the model is `unconfident' about (!). Hence for our problem self-reflection does not seem very accurate.

Another observations is that the distribution of answers varies quite a lot in Fig.~\ref{fig:likert_scale_palm}, as can be seen by the different proportions (widths) of colors within each bar. This is problematic, since it suggests that the task of self-reflection is less important than how the the self-reflection prompt is phrased.

Finally, we note that depending on how the questions are phrased, quite a few answers fall outside the expected scale (NA in brown, Fig.~\ref{fig:likert_scale_palm}); the \textit{neither agree nor disagree} was hard for the model to produce. However, when we made this option simpler by phrasing it as \textit{neutral} it also affected how many times other Likert scale points were selected (compare row 7 (C+N) with row 8 (C)).

\para{Binary responses.}
To obtain a binary confidence score, we use a prompt asking the LVLM about its confidence and requiring a yes/no response, \eg \texttt{\footnotesize{Are you confident that "A" correctly answers "Q"? Output yes or no}}. In preliminary results we found that the obtained binary scores do not correlate highly with the correctness of an answer (data not shown). Hence binary self-reflection did not work for us either. To understand why, we again test several paraphrases of the same prompt and plot the distribution of answers in Fig.~\ref{fig:prompt_variations}. Again, we see that the exact phrasing of the prompt seems to be the dominant factor in these results. For example, using the top question PaLI is confident of its answer in 65\% of the cases, whereas using the bottom question it is confident only in 17\% of the cases. For PaLM, results vary between 40\%-69\%, which is comparatively less but still a worrying amount of variation.

Overall, we conclude that the phrasing of the prompt dominates the actual task of self-reflection, which suggests that it may not be a reliable method for estimating confidences. 
\section{External Evaluation}\label{sec:external_eval}

Judging others is typically easier than judging yourself. So in this section we experiment with using an LVLM as external evaluator to select which of the nine models provides the best answer to a given question. This is similar to the evaluators in \cite{Hao2023ReasoningWL, shinn2023reflexion, Yao2023TreeOT}.
As external evaluator we use a large, instruction tuned version of PaLM, 
denoted as \texttt{PaLM 2-L-IT}.

\subsection{Choosing the Best Answer}

A straightforward approach is to simply provide the external evaluator with all possible answers and have it select one. We try this approach. Furthermore, we also use more elaborate prompts with intermediate questions, for example to match the entity type of the question and the answer, or to emphasise answers which occur multiple times. Finally, we try including the retrieved Wikipedia context to aid the choice.
However, preliminary experiments evaluating these three prompting strategies never achieve more than 43\% accuracy, significantly lower than the 49.4\% baseline given by PaLM with Lens (full results in Appendix \ref{bestAnswerChoice}).

\subsection{Evaluating the Reasoning Process}

\begin{figure}[!b]
    \hspace{-2.15cm}    
    \includegraphics[width=1.3\textwidth]{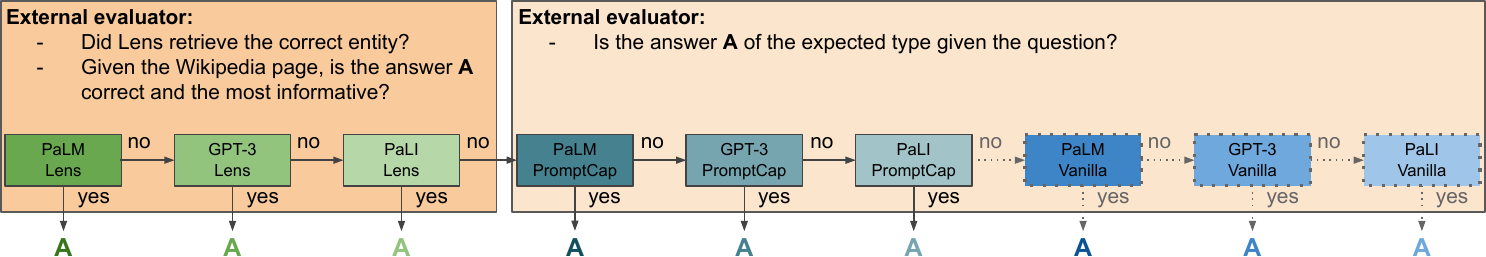}
    \caption{\small{LVLM cascade by evaluating the reasoning process using an external evaluator.}}
    \label{fig:cascade_diagram}
\end{figure}

\begin{table}[!t]
    \vspace{-.9cm}
    \caption{\small{Results of cascade and evaluator performance.}}
    \centering
    \begin{subtable}{0.48\linewidth}
        \centering
        \resizebox{!}{10mm}{
        \begin{tabular}{p{4.5cm}p{1.8cm}}
            \toprule
            \textbf{} & \textbf{{BEM}}              \\
            \cmidrule{2-2}
            {PaLM with Lens (baseline)} & {48.8\%} \\
            \midrule
            {Cascade with vanilla LVLMs} & {53.2\%} \\
            {Cascade without vanilla LVLMs} & \textbf{{53.4\%}} \\
            \bottomrule            
        \end{tabular}
        }
        \vspace{-2mm}
        \caption{Accuracy of LVLM cascade.}
        \label{results}
    \end{subtable}%
    ~
    \begin{subtable}{0.48\linewidth}
        \centering
        \resizebox{!}{10mm}{
        \begin{tabular}{lcc}
            \toprule
            \textbf{} & \textbf{{Precision}} & \textbf{{Recall}}              \\
            \cmidrule{2-3}
        {Lens} & {77.4\%} & {76.7\%} \\
        {PromptCap} & {29.1\%} & {86.3\%} \\
        {Vanilla} & {19.3\%} & {86.5\%} \\
        \bottomrule
        \end{tabular}
        }
        \vspace{-2mm}
        \caption{\small{Precision and recall of evaluator.}}
        \label{precisionRecall}
    \end{subtable}
    \vspace{-5mm}
\end{table}%

Since simple selection does not work, we resort to having the evaluator carefully analyse the reasoning process of each LVLM to judge  whether the answer is sound or not.

A Lens-based LVLM (1) queries Lens to get the Wikipedia Entity and then (2) extracts the answer from its corresponding Wikipedia page. We have the evaluator analyse the success of both steps using multi-step prompts. In particular, for (1) we first ask whether Lens gave any result at all. If it did, we have the evaluator extract the entity type of the question (\eg bird). Then we ask whether the retrieved entity (\eg sparrow, Matterhorn) is indeed a bird or not. For (2) we give the evaluator the question and the Wikipedia page. Then we have it list all possible answers to the question, select the most informative answer (\eg Central Europe is more informative than Europe), and compare that to the answer given by the Lens-based LVLM. If either of the two reasoning steps is judged unsuccessful, the answer of this LVLM is discarded. Full prompts are given in Appendix~\ref{evaluatorPrompts}.

For the PromptCap-based LVLMs we have a different strategy. PromptCap always gives a caption, so there are no failure modes there. Furthermore, the captions rarely explicitly contain the information asked for in the question, so we cannot verify that either. Instead, we have the evaluator compare via a two-step prompt whether the answer of the LVLM is of the correct information type given the question: \eg when asked about the name of a country we may expect Switzerland but not Europe. Since this strategy only relies on the question and the answer, we also use it for the vanilla LVLMs.

At this point, we have binary decisions for whether the answer for a single LVLM is sound. To combine everything into a single model, we use a cascade \cite{gamacascades} which is visualized in Fig.~\ref{fig:cascade_diagram}. We choose the cascade order roughly based on performance: we first cascade all Lens-based LVLMs, then the PromptCap-based LVLMs, and then optionally the vanilla LVLMs. Within these, we first cascade PaLM, then GPT-3, then PaLI.

\para{Results.}
Tab. \ref{results} shows that we \emph{finally} made a successful ensemble of our LVLMs: the cascade which excludes the vanilla LVLMs improves SOTA from from $48.8\%$ to $53.4\%$; a total absolute improvement of $4.6\%$.

\para{Analysis.}
To gain more insight into why our cascade works, we first look at how well our evaluator can judge the reasoning process. We do this in terms of precision (\ie how many of the answers judged to be correct by the evaluator are truly correct) and recall (\ie how many of the correct answers are also judged to be correct by the evaluator). This is shown in Tab.~\ref{precisionRecall}. As can be seen, precision and recall are both high for evaluating the reasoning process of the Lens-based LVLMs, but precision is rather low for the other LVLMs. Hence our increased performance mostly stems from a good judgement of the Lens-based LVLM reasoning process.

Diving deeper into this result, we find that in the vast majority of the cases an answer is rejected because Lens did not retrieve any entity (!). This suggests a much easier cascade: Use the Lens-based LVLM if Lens retrieved an entity, otherwise resort to PromptCap to at least have some information of what is depicted. We implement this using PaLM leading to a simple 2-step cascade, which yields 52.4\% BEM accuracy, close to the 53.4\% of our best performing cascade. Hence most of our improvements simply come from observing whether Lens retrieved anything at all.

Finally, since our evaluator is based on the larger \texttt{PaLM 2-L-IT}, we also compare our cascade to the \texttt{PaLM 2-L-IT} + Lens model. This results in a BEM accuracy of 52.7\%. Hence compared to this baseline, we achieve a much smaller performance improvement of 0.7\%.
\section{Conclusion}

We investigated \textit{how (not) to ensemble LVLMs for VQA} and found that an external evaluation model using the LVLMs in a cascade works best, increasing SOTA on EncyclopedicVQA from 48.8\% to 53.4\%. On the other hand, ablation on our cascade shows that a significant portion of the increase is given by (1) the larger size of our evaluator model, and (2) conditioning on whether Lens has identified a Wikipedia entity to decide which context should be added to the LVLM input.

There are additional limitations that derive from prompting LVLMs. Firstly, we only experimented with zero-shot prompts for extracting a confidence score via self-reflection. This is because, in our opinion, the inclusion of few-shot exemplars would risk biasing the model in this task. Additionally, the search space for natural language prompts is extremely large, and we do not claim to have carried out comprehensive prompt engineering.

Much of the large potential shown by the oracle experiment (a further 13.6\% above the new SOTA) still remains untapped. Future research can look into strategies to capture more of it. Overall, we have tried a wide range of methods and found that improving performance on Encyclopedic-VQA via ensembling is difficult, despite the large gain promised by the oracle.

\begin{ack}
We thank Andre Araujo and Vitto Ferrari for their useful insights and discussions through the project.
\end{ack}

{\small
\bibliographystyle{abbrv}
\bibliography{shortstrings,loco,loco_extra}
\clearpage}

\appendix
\section{Oracle experiment results}\label{oracleResultsAppendix}
In Tab.~\ref{tab:oracle_results_appendix} we show a more extensive overview of the baseline performance and the performance of oracle ensembles.

\begin{table}[!h]
  \caption{Oracle and Baseline results for EncyclopedicVQA. We show the results of the individual models (from~\cite{mensink2023encyclopedic}) and for different ensembles in the oracle setting. The three base models are PaLI, PaLM and GPT-3.
  The ensemble with all 9 models outperforms the best single model by $18.2\%$.
  \vspace{5pt}
  }
  \label{tab:oracle_results_appendix}
  \centering
  \begin{tabular}{lc}
    \toprule
    \textbf{Ensemble} & \textbf{Acc. (BEM)}              \\
    \midrule
    PaLI vanilla & $13.0\%$\\
    PaLI PromptCap & $17.8\%$\\
    PaLI Lens Section & $28.1\%$\\[2mm]    
    GPT-3 vanilla & $15.5\%$\\
    GPT-3 PromptCap & $25.6\%$\\
    GPT-3 Lens Section & $44.6\%$\\[2mm]
    PaLM vanilla & $19.7\%$\\
    PaLM PromptCap & $29.7\%$\\
    PaLM Lens Section & \textbf{\textit{48.8\%}}\\
    \midrule
    3 base models with PromptCap & $38.9\%$ \\
    3 base models with Lens & $56.6\%$   \\
    3 PaLM models (vanilla, PromptCap, Lens) & $58.8\%$\\[1mm]
    6 base models (all PromptCap and all Lens) & $65.8\%$ \\[1mm]
    9 base models (all PromptCap, Lens, and vanilla)& \textbf{67.0\%} \\
    \bottomrule
  \end{tabular}
\end{table}
\section{Majority voting results}\label{majorityVoting}

In Tab.~\ref{votingResults} we show the results of majority voting using exact matching and BEM-based soft matching.

 \begin{table}[htbp]
  \caption{Results of ensembling via majority voting. Exact matching is performed after normalisation via lower-casing and punctuation removal.\vspace{5pt}}
  \label{votingResults}
  \centering
  \begin{tabular}{lc}
    \toprule
    \textbf{} & \textbf{Acc. (BEM)}              \\
    \cmidrule{2-2}
    PaLM with Lens (baseline) & $48.8\%$ \\
    \midrule
    Majority voting via exact matching & $40.9\%$ \\
    Majority voting via BEM soft-matching & $45.3\%$ \\
    \bottomrule
  \end{tabular}
\end{table}
\section{Calibration}
\label{sec:calibration}

\subsection{Re-calibration with temperature scaling}
We use temperature scaling to re-calibrate the sequence probabilities for each of the LVLMs.
For each model we select the temperature $t^*$ with the lowest ECE / Brier score, simply by evaluating multiple temperature values.

\begin{figure}[tbh]
    \centering
    \includegraphics[width=.40\textwidth]{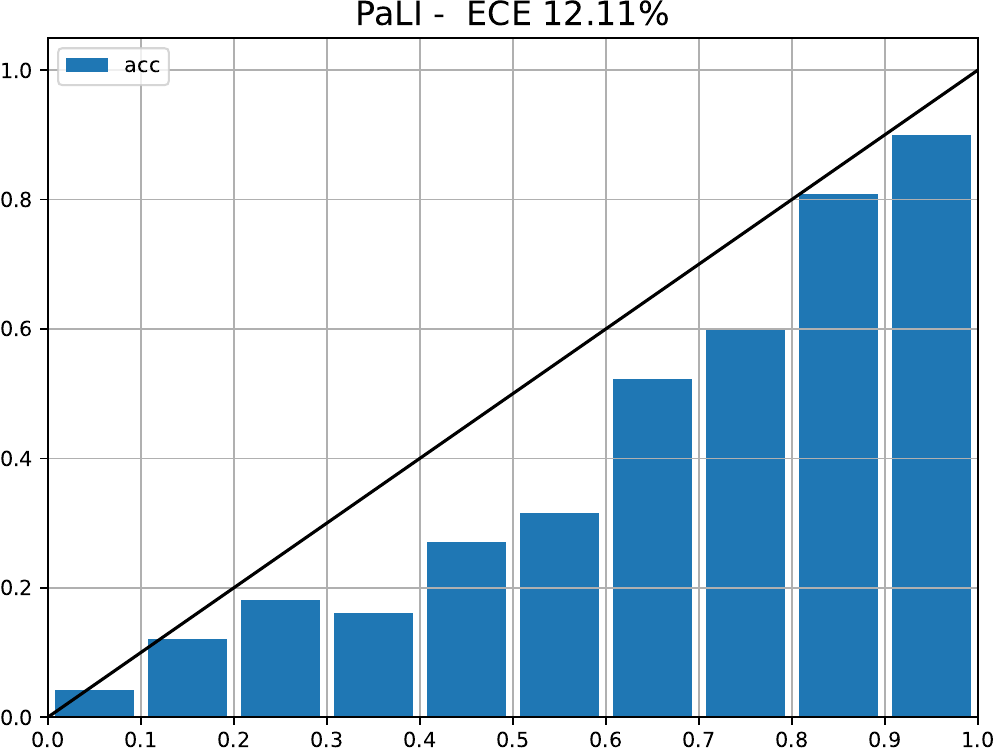}\quad
    \includegraphics[width=.40\textwidth]{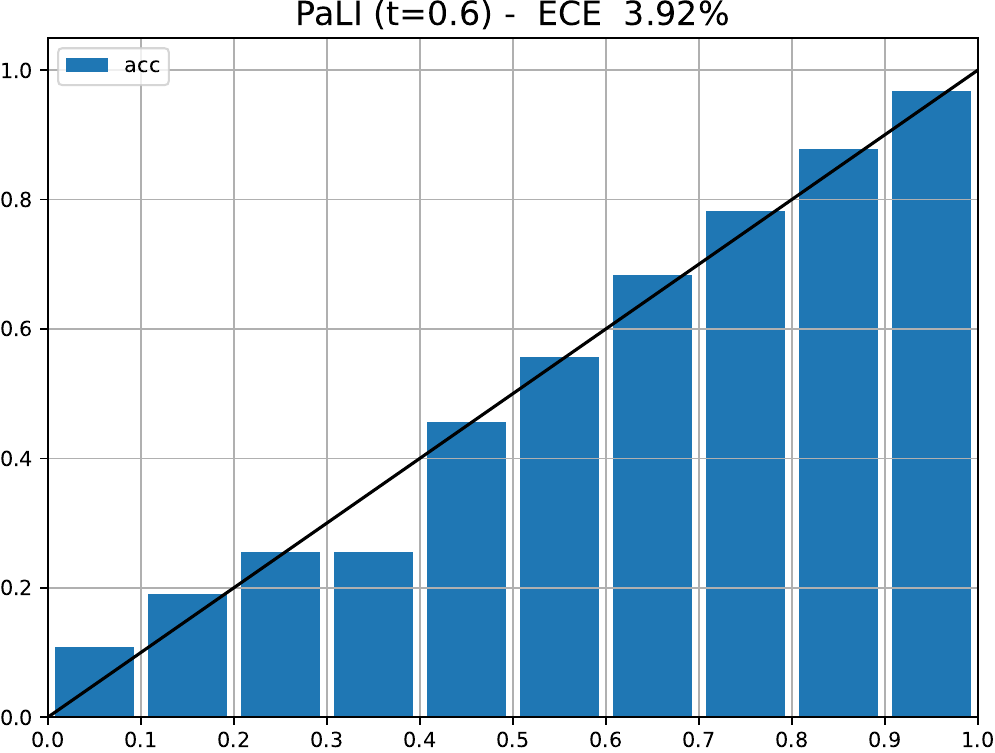} \vspace{3mm}\\
    \includegraphics[width=.40\textwidth]{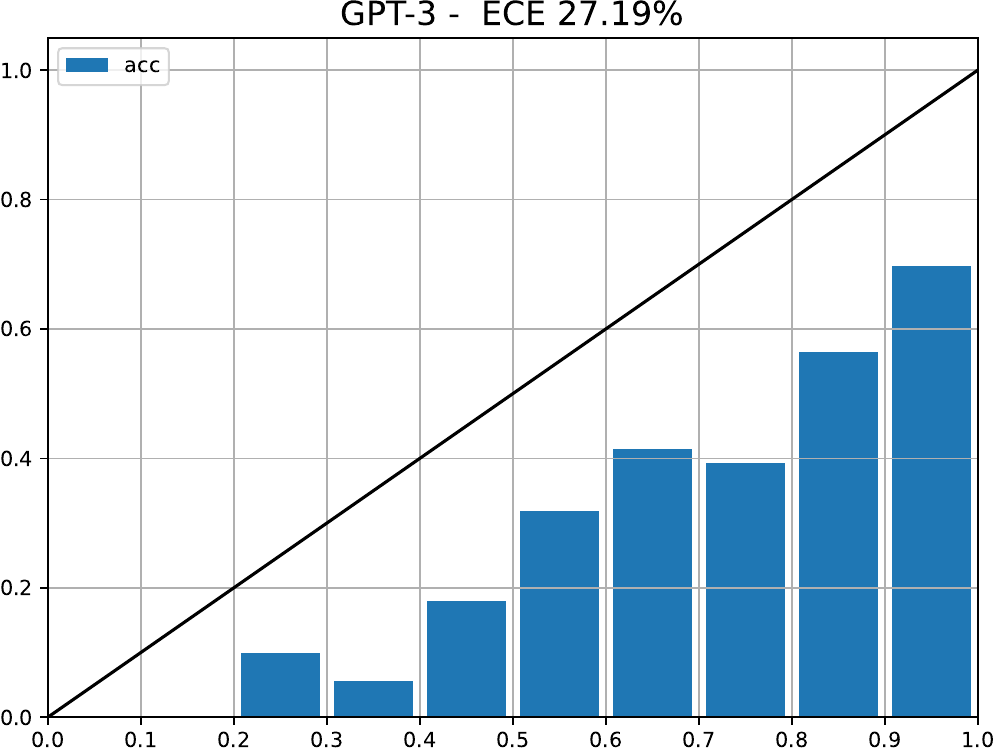}\quad
    \includegraphics[width=.40\textwidth]{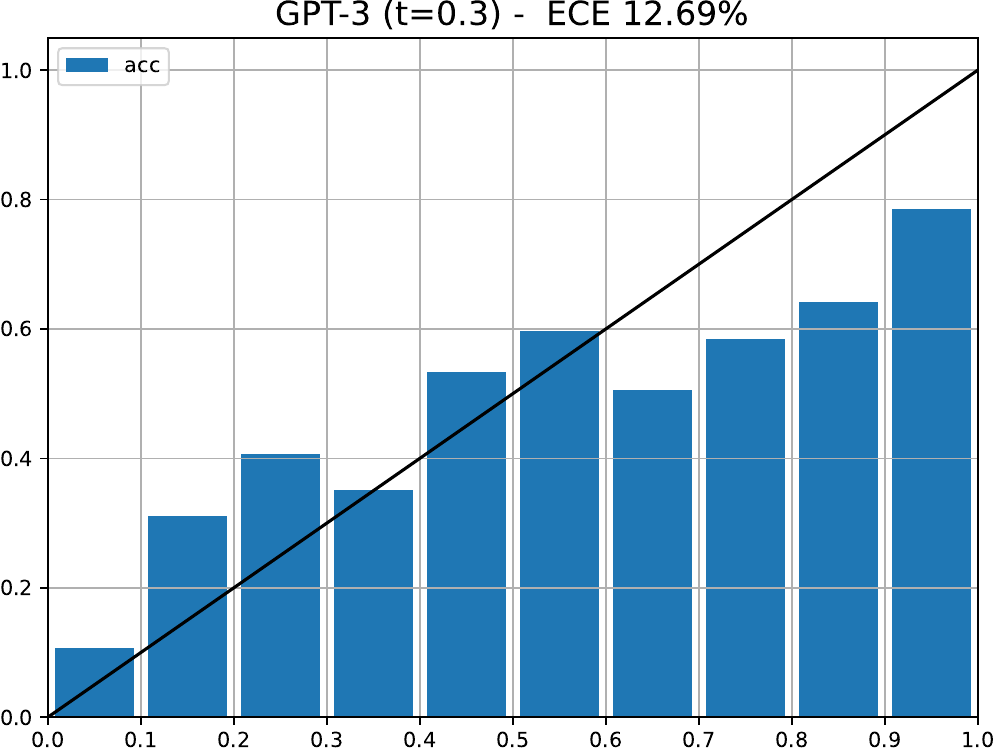} \vspace{3mm}\\
    \includegraphics[width=.40\textwidth]{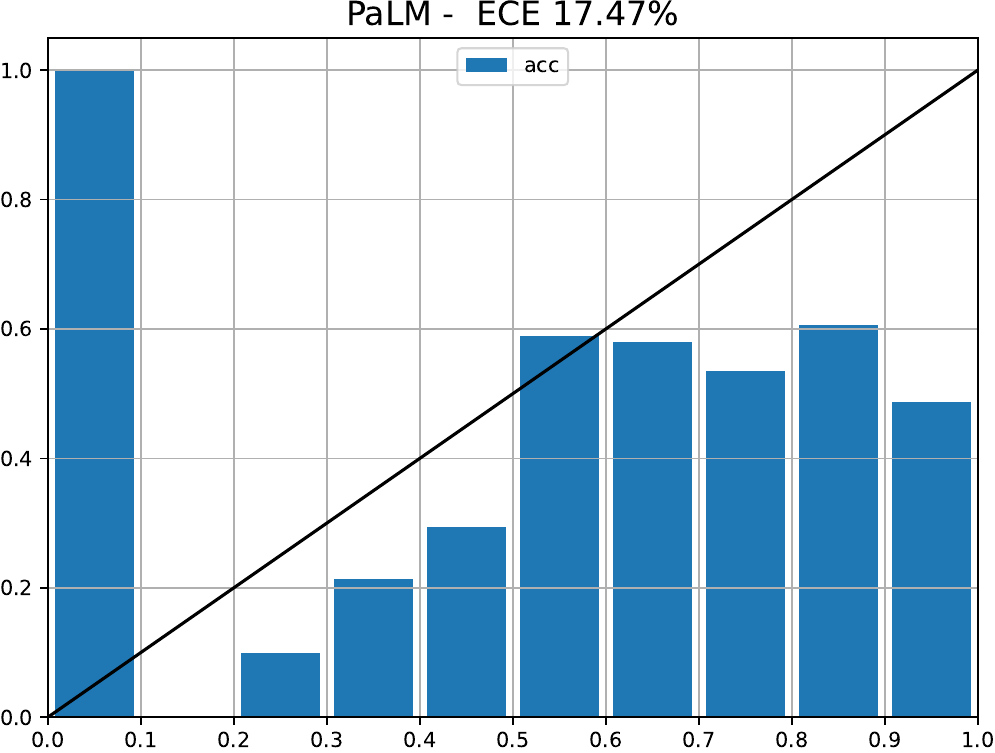}\quad
    \includegraphics[width=.40\textwidth]{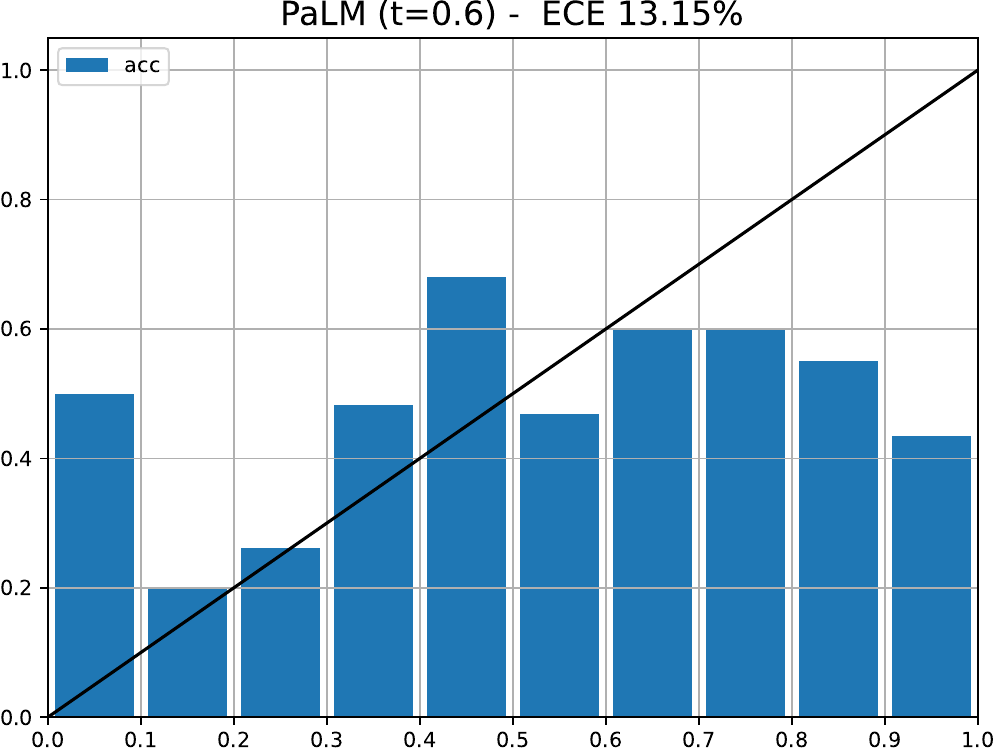}
    \caption{ECE diagrams for the three LVLMs before and after temperature scaling.}
    \label{fig:calibration_ece}
\end{figure}

ECE diagrams before and after temperature scaling are shown in Fig.~\ref{fig:calibration_ece}.

\subsection{Effect of length normalization on calibration}
Normalizing the sequence probabilities for length results in worse ECE and Brier scores than using unnormalized probabilities. This is true for all the Lens-based LVLMs, as shown in Tab.~\ref{calib}. However, applying recalibration via temperature scaling significantly reduces this gap, and even gives (slightly) better Brier scores for PaLI and GPT-3 after length normalization  (average difference in ECE: 0.068$\pm$0.028 without recalibration, 0.015$\pm$0.019 with recalibration; average difference in Brier score: 0.031$\pm$0.023 without recalibration, 0.01$\pm$0.027 with recalibration).
On the other hand, length normalization improves the BEM accuracy of all classical ensembling methods, as evidenced in Tab.~\ref{ensemblingResults}. We hence use normalized probabilities.

\subsection{Calibration in perspective}
To put these calibration results in perspective, we compare our observations and results with the ImageNet classification experiments in~\cite{minderer2021revisiting}. In~\cite{minderer2021revisiting}, 26 models trained on ImageNet-1K are evaluated for different calibration metrics. The authors observe correlation between the calibration metrics, and less of a correlation between calibration metrics and classification accuracy. Moreover, they report Brier scores in the range 17.6 - 58.2 (mean 29.9) and ECE scores in the range 1.4 - 8.4 (mean 3.7). 
The LVLMs on our task have ECE scores which are slightly higher, and Brier scores which are below their mean. 
Based on these results, we conclude that the sequence probabilities are (reasonably) well calibrated and we aim to compare these across the different models.

\begin{table}[htbp]
  \caption{Calibration results before and after normalizing the sequence probabilities for length.}
  \vspace{1.5mm}
  \label{calib}
  \centering
   \begin{tabular}{lccc|ccc}
        \toprule
        & \multicolumn{3}{c|}{\textbf{Before normalization}} & \multicolumn{3}{c}{\textbf{After normalization}} \\
        \cmidrule{2-7}
        & PaLI & PaLM & GPT-3 & PaLI & PaLM & GPT-3\\
        \cmidrule{2-7}
        \textbf{ECE} ($\downarrow$) & 0.081 & 0.107 & 0.177 & 0.121 & 0.175 & 0.272 \\
        \textbf{Brier} ($\downarrow$) & 0.174 & 0.219 & 0.246 & 0.182 & 0.272 & 0.278 \\
        \midrule
        \textbf{ECE}-re ($\downarrow$) & 0.037 & 0.095 & 0.121 & 0.039 & 0.131 & 0.127 \\
        \textbf{Brier}-re ($\downarrow$) & 0.167 & 0.215 & 0.229 & 0.164 & 0.256 & 0.221 \\
        \bottomrule
   \end{tabular}
\end{table}

\begin{table}[htbp]
  \caption{BEM accuracy of classical ensembling methods before and after normalizing the sequence probabilities for length. All results are on the Lens retrieval setup. Note that our logistic regression classification method one-vs-rest (OvR), taking the argmax of the three classifiers to output which LVLM to use.}
  \vspace{1.5mm}
  \label{ensemblingResults}
  \centering
   \begin{tabular}{lcc}
        \toprule
        & \textbf{Before normalization} & \textbf{After normalization} \\
        \cmidrule{2-3}
        Max probability & 48.7\% & 49.2\% \\
        Weighted voting & 48.6\% & 49.2\% \\
        Logistic regression & 49.9\% & 50.6\% \\
        \bottomrule
   \end{tabular}
\end{table}

\section{Prompt Variations}
\label{sec:prompt_variations}
\begin{figure}[!ht]
    \mbox{}\hspace{-19mm}{}\includegraphics[width=1.3\textwidth]{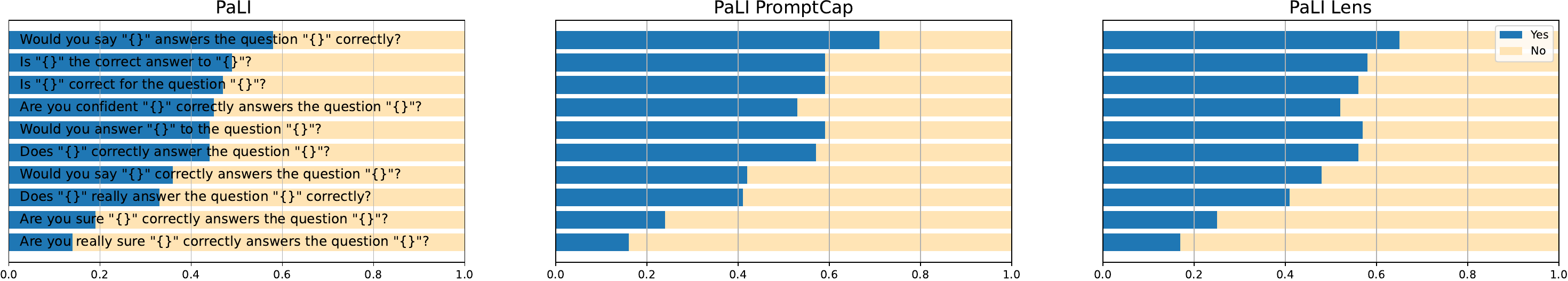}\vspace{3mm}\\
    \mbox{}\hspace{-19mm}{}\includegraphics[width=1.3\textwidth]{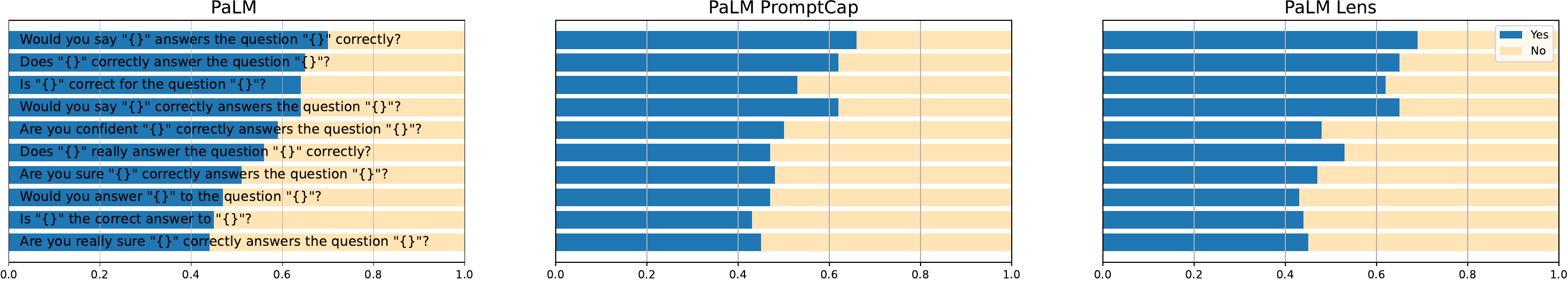}
    \caption{Prompt variations for PaLI and PaLM using the answers of three different base models. For each prompt the percentage of `yes' answers is shown.}
    \label{fig:prompt_variations_pali_and_palm}    
\end{figure}

In Fig.~\ref{fig:prompt_variations_pali_and_palm} we show the binary prompt variations for PaLI and PaLM for 10 different paraphrasings of the confidence prompt. We show the results for the three different base models (vanilla, PromptCap and Lens).
 
\clearpage
\section{Evaluator Model Prompts}\label{evaluatorPrompts}

All the exemplars in the following prompts are extracted from the Encyclopedic-VQA training set.

\subsection{Prompts for evaluating Lens LVLMs}\label{evaluatorPromptsLens}

\prompt{prompt:test_prompt_1}{Prompt to extract retrieval entity expected by the question.}{[ENTITY TYPE]: What type of entity does the [QUESTION] refer to?\newline\newline
[QUESTION]: What is one common name for this plant?\newline 
[ENTITY TYPE]: plant\newline\newline
[QUESTION]: What ants engage in interference competition against this ant colonies by 'plugging' entrances to the nest with sand and small rocks?\newline
[ENTITY TYPE]: ant\newline\newline
[QUESTION]: In what country is this amusement park located? \newline
[ENTITY TYPE]: amusement park\newline\newline
[QUESTION]: What does this bird try to do to the dingo by jumping into the air and kicking or stamping the dingo on its way down?\newline
[ENTITY TYPE]: bird\newline\newline
[QUESTION]: What kind of shopping centre was there before this one?\newline
[ENTITY TYPE]: shopping centre\newline\newline
[QUESTION]: How does this moth differ from the bride moths? \newline
[ENTITY TYPE]: moth\newline\newline
[QUESTION]: During the 1995 anniversary of ve day, what did seven heads of state take up at this hotel?\newline
[ENTITY TYPE]: hotel\newline\newline
[QUESTION]: What kind of structure is this memorial? \newline
[ENTITY TYPE]: memorial\newline\newline
[QUESTION]: What shape are the cones of this tree?\newline
[ENTITY TYPE]: tree\newline\newline
[QUESTION]: What was the nationality of the architect who presented plans for this theatre?\newline
[ENTITY TYPE]: theatre\newline\newline
[QUESTION]: What is the habitat of this animal?\newline
[ENTITY TYPE]: animal\newline\newline
[QUESTION]: Is this village accessible or inaccessible by road from east bergholt?\newline
[ENTITY TYPE]: village\newline\newline
[QUESTION]: What direction does the tunnel of this insect usually go?\newline
[ENTITY TYPE]: insect}

\clearpage

\promptcontinues{[QUESTION]: What does this church do with its biblical foundations and principles?\newline
[ENTITY TYPE]: church\newline\newline
[QUESTION]: Besides being the oldest, what other distinction does this pier have?\newline
[ENTITY TYPE]: pier\newline\newline
[QUESTION]: What do adult this butterflies eat?\newline
[ENTITY TYPE]: butterfly\newline\newline
[QUESTION]: What is depicted of king alexander jannaeus in this model?\newline
[ENTITY TYPE]: model\newline\newline
[QUESTION]: What sex of this dragonfly has a bright red abdomen?\newline
[ENTITY TYPE]: dragonfly\newline\newline
[QUESTION]: What is the climate like in this city?\newline
[ENTITY TYPE]: city\newline\newline
[QUESTION]: What does the queen of this wasp perform dance-like wing of?\newline
[ENTITY TYPE]: wasp\newline\newline
[QUESTION]: \{\}\newline
[ENTITY TYPE]:}

\prompt{prompt:test_prompt_2}{Prompt to check retrieved entity matches expected entity.}{Is the \{\} a \{\}? Answer just yes or no.}

\prompt{prompt:test_prompt_3}{Prompt to extract most informative answer from retrieved Wikipedia context.}{Answer the questions below. Only show your answer to [MOST INFORMATIVE ANSWER], do not show [ALL ANSWERS].\newline\newline
[ALL ANSWERS]: If there are more than one possible answer to the [QUESTION] that can be found in the [EVIDENCE], give all possible answers. If no answers can be found in the [EVIDENCE], say 'no answers'.\newline\newline
[MOST INFORMATIVE ANSWER]: Which answer is most informative among the ones above? Give that answer (hint: informative means containing the most precise and useful information). If you answered 'no answers' to [ALL ANSWERS], say 'no answer'.\newline\newline
[QUESTION]: \{\}\newline
[EVIDENCE]: \{\}\newline
[MOST INFORMATIVE ANSWER]:}

\clearpage

\prompt{prompt:test_prompt_4}{Prompt to assess equivalence of most informative answer and LVLM answer.}{[EQUIVALENT OR NOT]: Given the [QUESTION], the [RESPONSE] and the [MOST INFORMATIVE ANSWER], is the [RESPONSE] the same or equivalent to the [MOST INFORMATIVE ANSWER]? Just say yes or no.\newline\newline
[QUESTION]: Where has this plant become a naturalized archaeophyte?\newline
[RESPONSE]: Europe\newline
[MOST INFORMATIVE ANSWER]: Central Europe\newline
[EQUIVALENT OR NOT]: no\newline\newline
[QUESTION]: When do these birds defend fruit-bearing trees?\newline
[RESPONSE]: summertime\newline
[MOST INFORMATIVE ANSWER]: summer\newline
[EQUIVALENT OR NOT]: yes\newline\newline
[QUESTION]: What other trees does this tree grow with?\newline
[RESPONSE]: coniferous swamps\newline
[MOST INFORMATIVE ANSWER]: coniferous swamps\newline
[EQUIVALENT OR NOT]: yes\newline\newline
[QUESTION]: What might the aggressive reproductive characteristics of this tree be beneficially exploited for?\newline
[RESPONSE]: propagation\newline
[MOST INFORMATIVE ANSWER]: biofuel\newline
[EQUIVALENT OR NOT]: no\newline\newline
[QUESTION]: Is this tree a popular or unpopular choice for guitar soundboards?\newline
[RESPONSE]: Popular.\newline
[MOST INFORMATIVE ANSWER]: popular\newline
[EQUIVALENT OR NOT]: yes\newline\newline
[QUESTION]: When were the latest renovations or repairs in this park?\newline
[RESPONSE]: Recently\newline
[MOST INFORMATIVE ANSWER]: 2022\newline
[EQUIVALENT OR NOT]: no\newline\newline
[QUESTION]: Where does this bird depart from every march to april?\newline
[RESPONSE]: migration\newline
[MOST INFORMATIVE ANSWER]: North America\newline
[EQUIVALENT OR NOT]: no\newline\newline
[QUESTION]: How long do the leaves of this plant last?\newline
[RESPONSE]: year-round\newline
[MOST INFORMATIVE ANSWER]: throughout the year\newline
[EQUIVALENT OR NOT]: yes\newline\newline
[QUESTION]: When did henry ii begin spending money on this castle?\newline
[RESPONSE]: 1173\newline
[MOST INFORMATIVE ANSWER]: 1176\newline
[EQUIVALENT OR NOT]: no\newline\newline
[QUESTION]: Along with spain, where is this bird found?\newline
[RESPONSE]: australia and new zealand.\newline
[MOST INFORMATIVE ANSWER]: New Zealand.\newline
[EQUIVALENT OR NOT]: yes}

\clearpage

\promptcontinues{[QUESTION]: In meters, how long is this cathedral?\newline
[RESPONSE]: 120 m\newline
[MOST INFORMATIVE ANSWER]: 120\newline
[EQUIVALENT OR NOT]: yes\newline\newline
[QUESTION]: How many millimeters long are the glumes of this plant?\newline
[RESPONSE]: 2mm\newline
[MOST INFORMATIVE ANSWER]: 1.5-2.5\newline
[EQUIVALENT OR NOT]: yes\newline\newline
[QUESTION]: What is another predator of this ant besides insects and birds?\newline
[RESPONSE]: Mammals.\newline
[MOST INFORMATIVE ANSWER]: tarantulas\newline
[EQUIVALENT OR NOT]: no\newline\newline
[QUESTION]: When do this plant's flowers form fruit?\newline
[RESPONSE]: after pollination\newline
[MOST INFORMATIVE ANSWER]: after flowering\newline
[EQUIVALENT OR NOT]: no\newline\newline
[QUESTION]: When did hofwart de kirchheim reside at this castle?\newline
[RESPONSE]: 1200s\newline
[MOST INFORMATIVE ANSWER]: 13th century\newline
[EQUIVALENT OR NOT]: yes\newline\newline
[QUESTION]: When was the first record of this plant in ireland?\newline
[RESPONSE]: 11th century\newline
[MOST INFORMATIVE ANSWER]: 12th century\newline
[EQUIVALENT OR NOT]: no\newline\newline
[QUESTION]: What is the traditional use of this plant in medicine?\newline
[RESPONSE]: to stop bleeding.\newline
[MOST INFORMATIVE ANSWER]: stop blood loss\newline
[EQUIVALENT OR NOT]: yes\newline\newline
[QUESTION]: In addition to deltas and marshes, where does this bird feed?\newline
[RESPONSE]: estuaries.\newline
[MOST INFORMATIVE ANSWER]: wetlands\newline
[EQUIVALENT OR NOT]: no\newline\newline
[QUESTION]: What does this castle display about Richard Pennant?\newline
[RESPONSE]: Owned nearly 1,000 enslaved people.\newline
[MOST INFORMATIVE ANSWER]: slavery links\newline
[EQUIVALENT OR NOT]: yes\newline\newline
[QUESTION]: How often were rockets fired from this lighthouse starting in 1896?\newline
[RESPONSE]: Every year\newline
[MOST INFORMATIVE ANSWER]: annually\newline
[EQUIVALENT OR NOT]: yes\newline\newline
[QUESTION]: \{\}\newline
[RESPONSE]: \{\}\newline
[MOST INFORMATIVE ANSWER]: \{\}\newline
[EQUIVALENT OR NOT]:}

\clearpage
\newpage

\subsection{Prompts for evaluating PromptCap and Vanilla LVLMs}

\prompt{prompt:pcap_prompt_1}{Prompt to extract the type of information required by the question.}{[REQUIRED INFORMATION]: What information does the [QUESTION] ask for?\newline\newline
[QUESTION]: What is one common name for this plant?\newline
[REQUIRED INFORMATION]: common plant name\newline\newline
[QUESTION]: What ants engage in interference competition against this ant colonies by 'plugging' entrances to the nest with sand and small rocks?\newline
[REQUIRED INFORMATION]: ant species\newline\newline
[QUESTION]: In what country is this amusement park located?\newline
[REQUIRED INFORMATION]: name of country\newline\newline
[QUESTION]: What does this bird try to do to the dingo by jumping into the air and kicking or stamping the dingo on its way down?\newline
[REQUIRED INFORMATION]: purpose or action directed to the dingo\newline\newline
[QUESTION]: What kind of shopping centre was there before this one?\newline
[REQUIRED INFORMATION]: kind of shopping centre\newline\newline
[QUESTION]: How does this moth differ from the bride moths?\newline
[REQUIRED INFORMATION]: habit or feature a moth has\newline\newline
[QUESTION]: During the 1995 anniversary of ve day, what did seven heads of state take up at this hotel?\newline
[REQUIRED INFORMATION]: action taken up at hotel\newline\newline
[QUESTION]: What kind of structure is this memorial?\newline
[REQUIRED INFORMATION]: kind of structure\newline\newline
[QUESTION]: What shape are the cones of this tree?\newline
[REQUIRED INFORMATION]: shape\newline\newline
[QUESTION]: What was the nationality of the architect who presented plans for this theatre?\newline
[REQUIRED INFORMATION]: nationality\newline\newline
[QUESTION]: What is the habitat of this animal?\newline
[REQUIRED INFORMATION]: habitat of the animal\newline\newline
[QUESTION]: Is this village accessible or inaccessible by road from east bergholt?\newline
[REQUIRED INFORMATION]: yes/no answer\newline\newline
[QUESTION]: What direction does the tunnel of this insect usually go?\newline
[REQUIRED INFORMATION]: direction\newline\newline
[QUESTION]: What does this church do with its biblical foundations and principles?\newline
[REQUIRED INFORMATION]: action taken by church}

\clearpage

\promptcontinues{[QUESTION]: Besides being the oldest, what other distinction does this pier have?\newline
[REQUIRED INFORMATION]: unique feature of pier\newline\newline
[QUESTION]: What do adult this butterflies eat?\newline
[REQUIRED INFORMATION]: butterfly's prey or food\newline\newline
[QUESTION]: What is depicted of king alexander jannaeus in this model?\newline
[REQUIRED INFORMATION]: property or belonging\newline\newline
[QUESTION]: What sex of this dragonfly has a bright red abdomen?\newline
[REQUIRED INFORMATION]: sex of the animal\newline\newline
[QUESTION]: What is the climate like in this city?\newline
[REQUIRED INFORMATION]: type of climate\newline\newline
[QUESTION]: When was this landmark built?\newline
[REQUIRED INFORMATION]: year of construction\newline\newline
[QUESTION]: \{\}\newline
[REQUIRED INFORMATION]:}

\prompt{prompt:pcap_prompt_2}{Prompt to determine if the answer contains the required information.}{[GIVES REQUIRED INFORMATION]: Does the [ANSWER] to the [QUESTION] satisfy the [REQUIRED INFORMATION]?\newline\newline
[QUESTION]: What is one common name for this plant?\newline
[REQUIRED INFORMATION]: common plant name\newline
[ANSWER]: Peacock Flower\newline
[GIVES REQUIRED INFORMATION]: yes\newline\newline
[QUESTION]: What ants engage in interference competition against this ant colonies by 'plugging' entrances to the nest with sand and small rocks?\newline
[REQUIRED INFORMATION]: ant species\newline
[ANSWER]: ants\newline
[GIVES REQUIRED INFORMATION]: no\newline\newline
[QUESTION]: In what country is this amusement park located?\newline
[REQUIRED INFORMATION]: name of country\newline
[ANSWER]: austria\newline
[GIVES REQUIRED INFORMATION]: yes\newline\newline
[QUESTION]: What does this bird try to do to the dingo by jumping into the air and kicking or stamping the dingo on its way down?\newline
[REQUIRED INFORMATION]: purpose or action directed to the dingo\newline
[ANSWER]: scare it away\newline
[GIVES REQUIRED INFORMATION]: yes}

\clearpage

\promptcontinues{
[QUESTION]: What kind of shopping centre was there before this one?\newline
[REQUIRED INFORMATION]: kind of shopping centre\newline
[ANSWER]: Westfield shopping centre\newline
[GIVES REQUIRED INFORMATION]: no\newline\newline
[QUESTION]: How does this moth differ from the bride moths?\newline
[REQUIRED INFORMATION]: habit or feature a moth has\newline
[ANSWER]: diurnal\newline
[GIVES REQUIRED INFORMATION]: yes\newline\newline
[QUESTION]: During the 1995 anniversary of ve day, what did seven heads of state take up at this hotel?\newline
[REQUIRED INFORMATION]: action taken up at hotel\newline
[ANSWER]: Conrad London\newline
[GIVES REQUIRED INFORMATION]: no\newline\newline
[QUESTION]: What kind of structure is this memorial?\newline
[REQUIRED INFORMATION]: kind of structure\newline
[ANSWER]: obelisk\newline
[GIVES REQUIRED INFORMATION]: yes\newline\newline
[QUESTION]: What shape are the cones of this tree?\newline
[REQUIRED INFORMATION]: shape\newline
[ANSWER]: leaves and tendrils\newline\
[GIVES REQUIRED INFORMATION]: no\newline\newline
[QUESTION]: What was the nationality of the architect who presented plans for this theatre?\newline
[REQUIRED INFORMATION]: nationality\newline
[ANSWER]: Le Corbusier\newline
[GIVES REQUIRED INFORMATION]: no\newline\newline
[QUESTION]: What is the habitat of this animal?\newline
[REQUIRED INFORMATION]: habitat of the animal\newline
[ANSWER]: river deltas\newline
[GIVES REQUIRED INFORMATION]: yes\newline\newline
[QUESTION]: Is this village accessible or inaccessible by road from east bergholt?\newline
[REQUIRED INFORMATION]: yes/no answer\newline
[ANSWER]: \newline
[GIVES REQUIRED INFORMATION]: no\newline\newline
[QUESTION]: What direction does the tunnel of this insect usually go?\newline
[REQUIRED INFORMATION]: direction\newline
[ANSWER]: termites\newline
[GIVES REQUIRED INFORMATION]: no\newline\newline [QUESTION]: What does this church do with its biblical foundations and principles?\newline
[REQUIRED INFORMATION]: action taken by church\newline
[ANSWER]: people praying\newline
[GIVES REQUIRED INFORMATION]: no}

\clearpage

\promptcontinues{
[QUESTION]: Besides being the oldest, what other distinction does this pier have?\newline
[ENTITY TYPE]: unique feature of pier\newline
[ANSWER]: longest\newline
[GIVES REQUIRED INFORMATION]: yes\newline\newline
[QUESTION]: What do adult this butterflies eat?\newline
[REQUIRED INFORMATION]: butterfly's prey or food\newline
[ANSWER]: pollen\newline
[GIVES REQUIRED INFORMATION]: yes\newline\newline
[QUESTION]: What is depicted of king alexander jannaeus in this model?\newline
[REQUIRED INFORMATION]: human feature, property or belonging\newline
[ANSWER]: Jerusalem model\newline
[GIVES REQUIRED INFORMATION]: no\newline\newline
[QUESTION]: What sex of this dragonfly has a bright red abdomen?\newline
[REQUIRED INFORMATION]: sex\newline
[ANSWER]: male\newline
[GIVES REQUIRED INFORMATION]: yes\newline\newline
[QUESTION]: What is the climate like in this city?\newline
[REQUIRED INFORMATION]: type of climate\newline
[ANSWER]: New York\newline
[GIVES REQUIRED INFORMATION]: no\newline\newline
[QUESTION]: When was this landmark built?\newline
[ENTITY TYPE]: year of construction\newline
[ANSWER]: 1790\newline
[GIVES REQUIRED INFORMATION]: yes\newline\newline
[QUESTION]: \{\}\newline
[REQUIRED INFORMATION]: \{\}\newline
[ANSWER]: \{\}\newline
[GIVES REQUIRED INFORMATION]:}

\clearpage
\section{Answer choice results}\label{bestAnswerChoice}

In Tab. \ref{tab:multichoice} we show the results of choosing the best answer among those given by the LVLMs, using different prompts.

\begin{table}[!h]
  \caption{Results of choosing the best answer via prompting, all well below the baseline (48.8\%). When \texttt{\footnotesize{[EXEMPLARS]}} are part of the prompt, these are the same as in Appendix \ref{evaluatorPromptsLens} Prompt~\ref{prompt:test_prompt_1}.\vspace{5pt}}
  \label{choosingResults}
  \centering
  \begin{tabular}{p{1.7cm}p{9.2cm}p{1.8cm}}
    \toprule
    \textbf{Sample size} & \textbf{Prompt} & \textbf{Acc. (BEM)}           \\
    \midrule
    {} & \texttt{\scriptsize{Consider the question [QUESTION] and the list with possible answers [ALL ANSWERS].\newline
    [FINAL ANSWER]: Among the answers in [ALL ANSWERS], which one is the most likely to be correct?\newline Hint: Repeated answers are more likely to be correct.}} & $40.0\%$ \\
    \cmidrule{2-3}
    100 & \texttt{\scriptsize{Consider the question [QUESTION], the answer [ANSWER] and the list with possible answers [POSSIBLE ANSWERS].\newline
    [SIMILAR ANSWERS]: Count the number of answers in [POSSIBLE ANSWERS] which are identical or semantically similar to [ANSWER].\newline
    If the majority of the answers are similar, then [ANSWER] is the [CORRECT ANSWER].\newline
    If not, consider the following:\newline
    [BETTER ANSWER]: Is there an answer in [POSSIBLE ANSWERS] which answers the question [QUESTION] more precisely and concisely than [ANSWER]?\newline
    If yes, then this answer is the [CORRECT ANSWER].\newline
    If not, use [ANSWER] as the [CORRECT ANSWER].}} & $43.0\%$ \\
    \midrule
    {} & \texttt{\scriptsize{(1)\newline [ENTITY TYPE]: Based on the [QUESTION], what kind of entity would you expect the answer to be?\newline
    [EXEMPLARS]\newline\newline
    (2)\newline[BEST ANSWER]: Given the [QUESTION], the comma-separated [ANSWERS] and the [ENTITY TYPE], which of the [ANSWERS] is the best and most informative for the question while also matching the [ENTITY TYPE]? You must choose only one of the comma-separated [ANSWERS] and output it without changes.}} & $40.0\%$ \\
    \cmidrule{2-3}
    500 & \texttt{\scriptsize{(1)\newline Answer the questions below. Only show your answer to [MOST INFORMATIVE ANSWER], do not show [ALL ANSWERS].\newline
    [ALL ANSWERS]: If there are more than one possible answer to the [QUESTION] that can be found in the [EVIDENCE], give all possible answers. If no answer can be found in the [EVIDENCE], say ‘no answers'.\newline
    [MOST INFORMATIVE ANSWER]: Which answer is most informative among the ones above? Give that answer (hint: informative means containing the most precise and useful information).If you answered ‘no answers' to [ALL ANSWERS], output ‘no answer'.\newline\newline
    (2)\newline [CHOICE]: Given the [QUESTION], the comma-separated [RESPONSES] and the [MOST INFORMATIVE ANSWER], which of the [RESPONSES] is most equivalent to the [MOST INFORMATIVE ANSWER]? Only return that response.}} & $32.6\%$ \\
    \bottomrule
  \end{tabular}
  \label{tab:multichoice}
\end{table}

\end{document}